%% file: main.tex
\pdfoutput=1

\documentclass[11pt]{article}

\usepackage{acl}
\usepackage{times}
\usepackage{amsmath} 
\usepackage{array}
\usepackage{enumitem}
\usepackage{setspace}

\usepackage{adjustbox}
\usepackage{algorithm}
\usepackage{algorithmic}
\usepackage{arydshln} 
\usepackage{float}
\usepackage{latexsym}
\usepackage{graphicx} 
\usepackage[T1]{fontenc}

\usepackage[utf8]{inputenc}

\usepackage{microtype}

\usepackage{booktabs}       
\usepackage{tabularx}
\usepackage{array}

\title{
CoAct: A Global-Local Hierarchy for Autonomous Agent Collaboration
}

\author{
Xinming Hou$^{1,2}$\thanks{\ \ This work was done during Xinming Hou's internship at Tencent AI Lab.}  \quad Mingming Yang$^2$ \quad Wenxiang Jiao$^{2}$\thanks{\ \ Wenxiang Jiao and Xing Wang are co‐corresponding authors.} \quad  Xing Wang$^{2}$\footnotemark[2] \\ 
\bf Zhaopeng Tu$^{2}$ \quad Wayne Xin Zhao$^{1}$\\
$^{1}$Gaoling School of Artificial Intelligence, Renmin University of China \\ $^{2}$Tencent AI Lab \\
\texttt{sherman.hou@gmail.com}\quad \texttt{\{joelwxjiao,brightxwang\}@tencent.com}
}

\begin{document}
\maketitle

\input{latex/0-abstract}

\input{latex/1-introduction}

\input{latex/2-method}
\input{latex/3-experiment}

\input{latex/4-conclusion}
\input{latex/5-limitations}
\input{latex/6-ethics}
\bibliography{ref}

\clearpage
\appendix

\section{Algorithm}
In this section, we present the algorithmic framework that underlies the CoAct design. The pseudocode for the algorithm is included below, providing a concise representation of the steps involved.

\input{latex/algorithm}

\section{Task Categorization}
\label{app:task-categ}
We categorize the examples that cannot be addressed by both ReAct and our CoAct into the \textbf{Hard} class. For those examples that only require one-step processing, we categorize them into the \textbf{Easy} class. The rest of examples are categorized into \textbf{Medium} class. As a result, the proportion of examples becomes 30\%:50\%:20\% for Easy:Medium:Hard. Table~\ref{tab:task_difficulty} shows the details.

\begin{table*}[t]
    \centering
    \begin{adjustbox}{max width=0.99\textwidth}
        \begin{tabular}{p{6cm}p{6.6cm}p{6.5cm}}
            \toprule
            \multicolumn{1}{c}{\bf Easy (30\%)} & \multicolumn{1}{c}{\bf Medium (\%50)} & \multicolumn{1}{c}{\bf Hard (20\%)} \\
            \midrule
            {\bf1. Search and Display} (10\%) & {\bf1. Product Information Retrieval} (20\%) & {\bf1. Advanced Product Selection} (6\%) \\
            - Search for ``\{\{keyword\}\}'' & - Show the least expensive {{product}} with a minimum storage capacity of \{\{min storage\}\}. & - Buy the best rating product from ``\{\{category\}\}'' category with at least 5 reviews and the product is least expensive \\
            - Show the ``\{\{product\}\}'' listings by \{\{sorting order\}\} & - Provide me with the full names of Bluetooth headphones from Sony, and also share the price range for the available models & - I am doing a market survey for one-stop market, show me the most expensive product from \{\{product category\}\} category \\
            - Show the most recent \{\{status\}\} order & - List products from \{\{product category\}\} category by \{\{order\}\} price & - Buy the highest rated product from the \{\{product category\}\} category within a budget \{\{dollar value\}\}. \\
            \hdashline\\[1pt]
            {\bf2. User Account and Profile} (10\%) & {\bf2. Review Handling} (15\%) & {\bf2. Data Analysis and Calculation} (7\%) \\
            - Subscribe to the newsletter of OneStopMarket & - Summarize customer reviews for \{\{product\}\}. & - How much refund I should expect from my order canceled in \{\{time\}\} if I cannot get the shipping fee refunded? \\
            - Today is 6/12/2023. Tell me how many fulfilled orders I have \{\{period\}\}, and the total amount of money I spent. & - List out reviewers, if exist, who mention about \{\{description\}\} & - What is the price range for products from \{\{brand\}\}? \\
            - I recently moved, my address is \{\{address\}\}, update my information on OneStop-Shopping accordingly & - Rate my recent purchase of {{product}} with {{num star}} stars, using my nickname \{\{nickname\}\}? & - What is the price range for products from \{\{product category\}\}? \\
            \hdashline\\[1pt]
            {\bf3. Wishlist and Product Actions} (5\%) & {\bf3. User Communication and Updates} (10\%) & {\bf3. Customer Interaction and Sentiment Analysis} (4\%) \\
            - Add this product to my wishlist & - Change the delivery address for my most recent order to \{\{address\}\} & - What do customers say about \{\{product type\}\} from \{\{manufature\}\} \\
            - Add \{\{product\}\} to my wish list. & - Draft an email to the shop owner via their contact us function for a coupon as \{\{reason\}\} & - List the customer names who thinks EYZUTAK phone cases are of good looking \\
            - Add a \{\{product\}\} to my wish list. & & Who gave \{\{stars\}\} for phone cases from EYZUTAK \\
            \hdashline\\[1pt]
            {\bf4. Basic Product Information} (5\%) & {\bf4. Order Management} (5\%) & {\bf4. Dynamic Data Handling} (3\%) \\
            - What is the price range of \{\{product\}\} in the One Stop Market? & - How much I spend \{\{time\}\} on shopping at One Stop Market? & - Tell me the status of my latest order and when will it arrive \\
            & - What is the total cost of my latest \{\{status\}\} order? & - Show me the ``\{\{product\}\}'' listings by \{\{sorting order\}\}. \\
            & - Get the order number of my most recent \{\{status\}\} order & - Show me products under \$\{\{price\}\} in ``\{\{product category\}\}'' category \\
            \bottomrule
        \end{tabular}
    \end{adjustbox}
    \caption{Analysis of task difficulty for the Shop task.}
    \label{tab:task_difficulty}
\end{table*}

\section{Prompts}
In this section, we provide a detailed presentation of prompts for two agents.

\input{latex/global-prompt}

\input{latex/local-prompt}

\end{document}

%% file: latex/0-abstract.tex
\begin{abstract}

Existing LLMs exhibit remarkable performance on various NLP tasks, but still struggle with complex real-world tasks, even equipped with advanced strategies like CoT and ReAct. In this work, we propose the CoAct framework, which transfers the hierarchical planning and collaboration patterns in human society to LLM systems. Specifically, our CoAct framework involves two agents: (1) A global planning agent, to comprehend the problem scope, formulate macro-level plans and provide detailed sub-task descriptions to local execution agents, which serves as the initial rendition of a global plan. (2) A local execution agent, to operate within the multi-tier task execution structure, focusing on detailed execution and implementation of specific tasks within the global plan.
Experimental results on the WebArena benchmark show that CoAct can re-arrange the process trajectory when facing failures, and achieves superior performance over baseline methods on long-horizon web tasks. 
Code is available at \url{https://github.com/xmhou2002/CoAct}.
\end{abstract}

%% file: latex/1-introduction.tex
\section{Introduction}

The field of artificial intelligence is progressively concentrating on uncovering innovative approaches for developing systems endowed with autonomy and self-adjustment capabilities. These features enable AI systems to manage increasingly complex, real-world natural language processing (NLP) tasks proficiently. To achieve successful outcomes, such systems necessitate robust planning and reasoning capabilities, as well as the capacity to adapt to errors and uncertainties.

While existing large language models~(LLMs) exhibit remarkable performance on a variety of NLP tasks, they still struggle with these complex reasoning tasks, encouraging the emergence of several strategies such as CoT~\cite{cot}, ReAct~\cite{react}, and self-refine~\cite{self-refine}. 
Despite these advancements, current explorations predominantly focus on a single LLM and a single memory stream. Recent studies~\cite{survey} indicate that the performance of a single LLM is constrained by the finite nature of the attention mechanism and hierarchical capacity, implying that there is potential for further improvement in autonomously handling real-world tasks. Consequently, this has led to the incorporation of the multi-agent collaboration concept, which has been extensively studied in the context of reinforcement learning~\cite{canese2021multi}, into the realm of LLMs research~\cite{guo2024large}.

In this paper, we propose the \textbf{CoAct} framework, which transfers the hierarchical planning and collaboration patterns in human society to LLM systems.
As we are building AI systems, it is natural to integrate human cognitive abilities into the development process, which has been widely followed in recent studies~\cite{laser,liang2023encouraging,qian2023communicative,he2024exploring}.
Specifically, our CoAct framework involves two agents: (1) \textit{Global planning agent}, to comprehend the problem scope, formulate macro-level plans and provide detailed sub-task descriptions to local execution agents, which serves as the initial rendition of a global plan. (2) \textit{Local execution agent}, to operate within the multi-tier task execution structure, focusing on detailed execution and implementation of specific tasks within the global plan. 
We expect this hierarchical planning framework to better understand the problem and solve it more accurately.
Experimental results on the WebArena benchmark show that
the proposed CoAct can re-arrange the process trajectory when facing failures, and achieves superior performance over ReAct on the real-world tasks.

\begin{figure*}[htb]
\centering
    \includegraphics[scale=0.38]{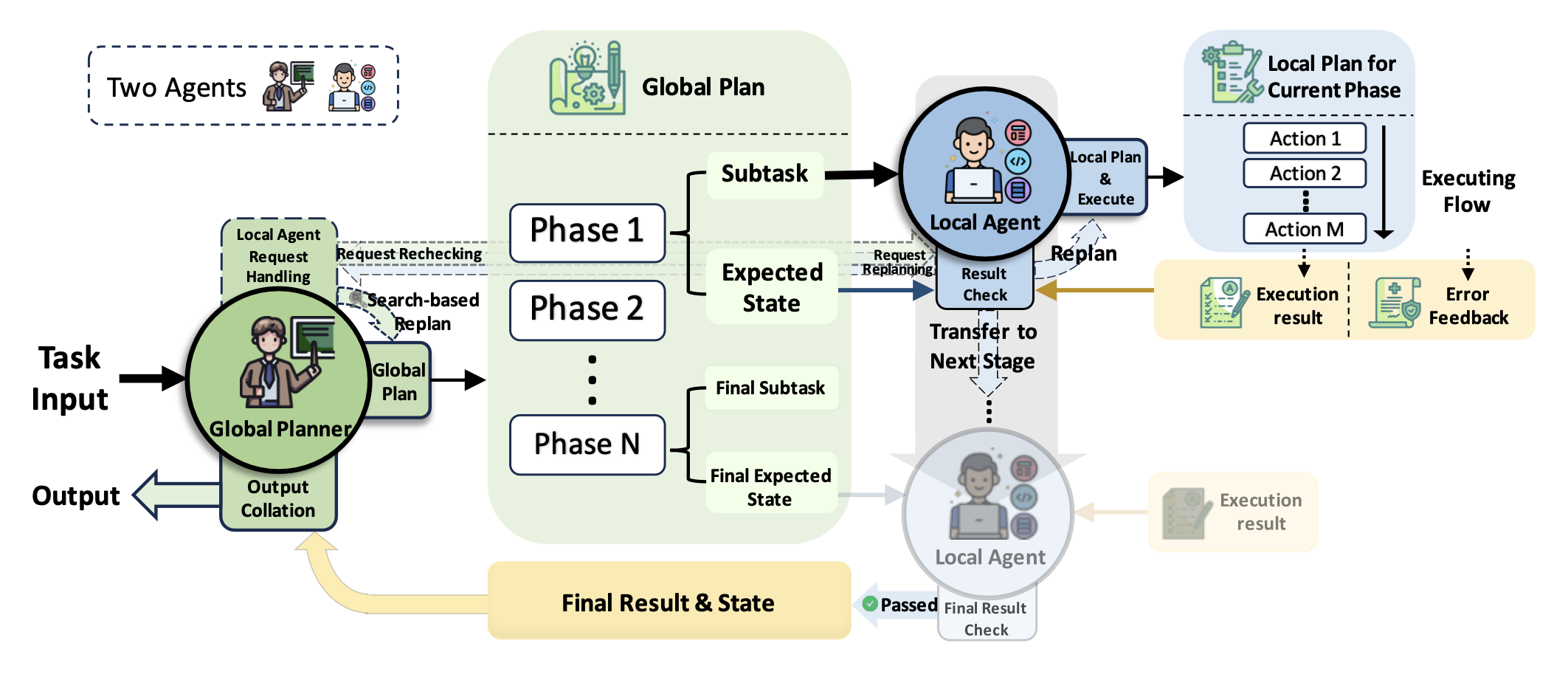}
    \caption{
    The framework of CoAct, which involves a global planning agent and a local execution agent to work together in a hierarchical relationship to accomplish tasks.
    }
    \label{fig:agent}
    \vspace{-5pt}
\end{figure*}

We summarize our key contributions as follows:
\begin{itemize}[noitemsep]
\item We introduced CoAct, a novel hierarchical planning framework that can enhance the reasoning ability of LLMs.
\item We empirically validated the effectiveness of CoAct on WebArena across diverse website environments.
\item We conducted extensive analysis of CoAct, providing insights in where it improves and how it can be further improved.
\end{itemize}

%% file: latex/2-method.tex
\section{Framework}

CoAct is an LLM-based multi-agent system designed for hierarchical collaboration among diverse agents. Figure~\ref{fig:agent} shows the framework, which includes decomposing tasks, assigning and communicating subtasks, analyzing and executing subtasks, collecting feedback, evaluating progress, and re-planning if necessary. Specifically:
\begin{itemize}
    \item The global planning agent decomposes tasks into subtasks (``Phase 1, Phase 2 ... Phase N'') and assigns them to the local execution agent.
    \item The local execution agent then analyzes and executes these subtasks while systematically collecting feedback (``Execution result'' and ``Error Feedback''). If execution falters, the agents re-plan to ensure success.
\end{itemize}

\subsection{Global Planning Agent}
\label{subsec:global}

\begin{figure}[!h]
\centering
    \includegraphics[scale=0.38]{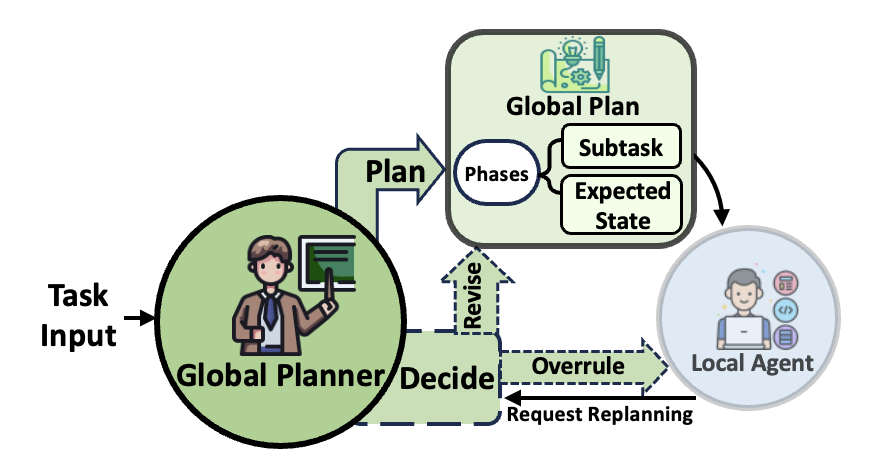}
    \caption{Workflow of global planning agent.}
    \label{fig:global}
    \vspace{-5pt}
\end{figure}

\noindent The global planning agent is crucial for navigating complex tasks. It starts by constructing comprehensive plans, dividing them into phased subtasks with clear outcomes. This agent manages the overall plan, ensuring each phase is well-defined. Upon requests from the local execution agent, the global planning agent reviews and decides on potential replanning, providing guidance and adjustments. It maintains the integrity of the global plan, suggesting modifications when necessary, and ensures the final task output aligns with the initial strategy.

\subsection{Local Execution Agent}
\label{subsec:local}

\begin{figure}[!h]
\centering
    \includegraphics[scale=0.38]{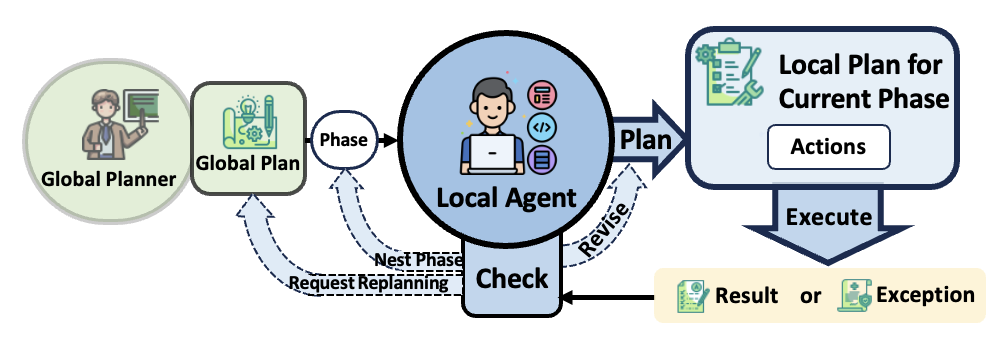}
    \caption{Workflow of local execution agent.}
    \label{fig:local}
    \vspace{-5pt}
\end{figure}

\noindent The local execution agent focuses on implementing specific subtasks within the global plan. This agent handles task execution, navigates web-based tasks, and ensures adherence to the overall strategy. It meticulously dissects each subtask, executes sequential actions, and verifies these actions against the global plan. The local execution agent evaluates progress based on collected feedback, deciding whether to revise its plan, request a new global plan, or proceed to the next phase. Detailed reporting of execution results is essential for ensuring alignment with the global objectives and providing a comprehensive summary of actions and outcomes.

%% file: latex/3-experiment.tex
\section{Experiment}

\begin{table}[t]
    \setlength{\tabcolsep}{3pt}
    \centering
    \begin{adjustbox}{max width=1.0\columnwidth} 
        \begin{tabular}{lrrrrrr}
            \toprule
            \textbf{Method} & \textbf{Shop} & \textbf{CMS} & \textbf{Reddit} & \textbf{Gitlab} & \textbf{Map} &
            \textbf{Avg} \\
            \midrule
            \textsc{Human} & -/- & -/- & -/- & -/- & -/- & 78.2 \\
            \hdashline
            \textsc{ReAct} & 12.0 & 11.0 & 9.0 & 7.0 & 8.0 & 9.4 \\
            \textsc{CoAct} & 22.0 & 14.0 & 12.0 & 9.0 & 12.0 & 13.8 \\
            \textsc{CoAct w/ FS} & \bf 24.0 & \bf 17.0 & \bf 14.0 & \bf 10.0 & \bf 15.0 & \bf 16.0 \\
            \bottomrule
        \end{tabular}
    \end{adjustbox}
    \caption{
    Performance of CoAct measured by task success rate~(SR) across five sub-tasks in WebArena. \textsc{Human} means human results from \cite{zhou2023webarena}; \textsc{CoAct w/ FS} denotes CoAct with force stop intervention.}
    \vspace{-5pt}
    \label{tab:main_results}
\end{table}

\subsection{Setting}

\paragraph{Models.}

We follow \citet{zhou2023webarena} to adopt ReAct~\cite{react} as the baseline, which asks the model to first perform CoT~\cite{cot} reasoning steps in the text before the action prediction. For our approach, we present two variants, i.e., CoAct and CoAct w/ FS, where FS denotes force stop intervention which forcibly terminates the dialogue when it exceeds a specified number of exchanges. 
We implement all the approaches based on the code released by \citet{zhou2023webarena}, and use~\texttt{gpt-3.5-turbo-16K-0613} as the backbone LLM. By default, we set the temperature to 1 to encourage the exploration.

\paragraph{Dataset.}

We evaluate our approach on WebArena~\cite{zhou2023webarena} dataset, which covers various tasks, namely, Shop, CMS, Reddit, Gitlab, and Map. It is a self-contained web environment crafted for developing autonomous agents, which generates websites across four distinct categories, faithfully replicating the functionality and data found in real-world counterparts.
The main challenges of WebArena are two-fold: 1) Observation bias, where LLMs fixate on the first piece of information they encounter without verifying its accuracy. 2) Action repetition, where failures in observation interpretation often makes LLMs repeat actions unnecessarily and ignore previously completed steps.
These challenges prevent models from accurately and efficiently performing complex web-based tasks. We sample 100 examples randomly from each task for experiments to ensure comprehensive coverage and representative evaluation. 
We report the success rate~(SR), i.e., the accuracy of task completion, to measure the performance of different approaches.

\subsection{Main Results}

\paragraph{CoAct achieves superior performance over ReAct on the real-world tasks.}

Table~\ref{tab:main_results} lists the results of ReAct and our CoAct on the WebArena benchmark. As seen, ReAct achieves 9.4\% success rate on average, which is comparable with that reported in \cite{zhou2023webarena}, i.e., 8.7\%, demonstrating the reasonableness of our implementation.
As for our CoAct, it improves ReAct by over 40\% success rate, and up to 70\% when with force stop intervention.
Specifically, CoAct can outperform ReAct across all the five tasks consistently, especially on the Shop task, suggesting its effectiveness and flexibility in solving real-world tasks.

\begin{figure}[t]
\centering
    \includegraphics[scale=0.45]{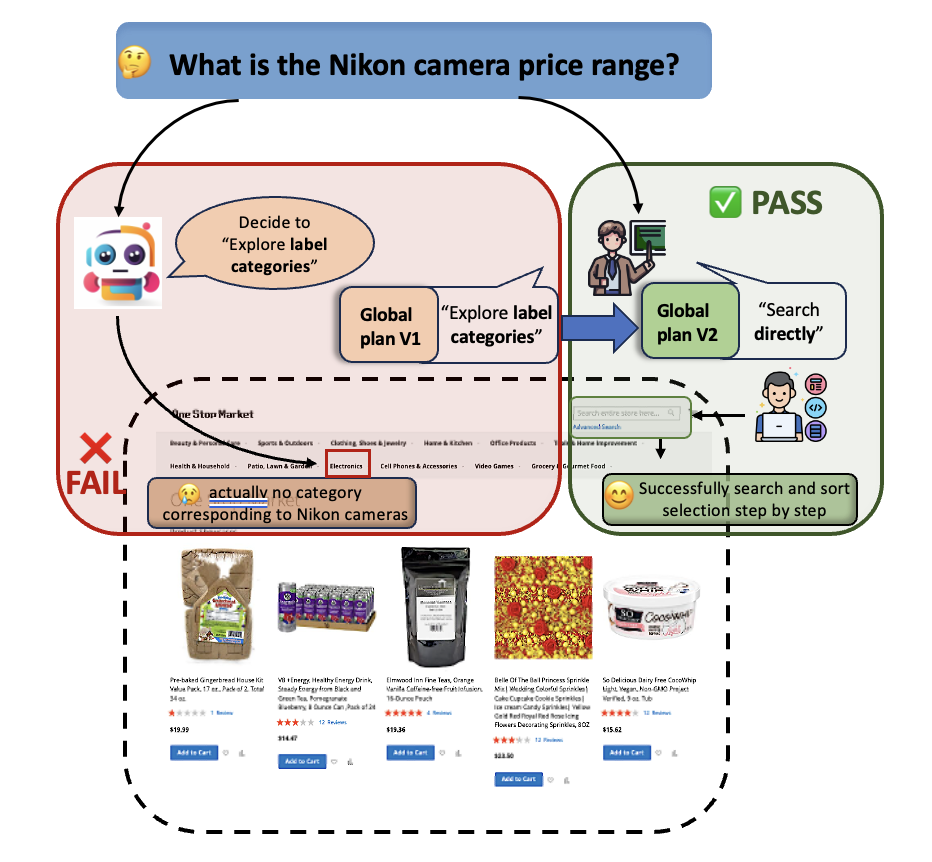}
    \caption{
    An example in the Shop task to show the advantage of CoAct over ReAct.}

    \label{fig:example}
\end{figure}

\paragraph{CoAct can re-arrange the process trajectory when facing failures.}

To gain a deeper understanding in where CoAct improves ReAct, we investigate the examples in Shop and present one in Figure~\ref{fig:example}.
In a basic ReAct setup, the agent follows a multi-step process: 1) identifying suitable subcategories, 2) locating the correct category, 3) sorting products within that category based on price, and 4) sequentially paging through to find the target item. However, when the tasks do not align with predefined categories, ReAct will struggle to address them as the agent accumulates excessive context information during category-seeking, preventing the model from recognizing the need to break out of the category search process after a failure.

However, our CoAct framework can well adapt to such scenarios. In CoAct, a global planning agent naturally segments the task execution process at a macro level, conveying sub-tasks to local execution agents. Despite accumulating context, prompts associated with sub-task descriptions guide redirection in case of planning errors. Local execution agents can request adjustments to the global plan, enabling macro-level re-planning.
Therefore, the core difference between CoAct and ReAct lies in context partitioning, attention allocation, and memory management. CoAct is more explicit, flexible, and universally applicable to solve real-world tasks across categories.

\subsection{Task Analysis}

\paragraph{Task categorization.}

While our CoAct outperforms ReAct significantly, it is still much worse than human performance. Therefore, it is necessary to understand the difficulty of different tasks, so as to develop strategies to further enhance the models. For simplicity, we conduct analysis on the Shop task by manual examination. We categorize the examples that cannot be addressed by both ReAct and our CoAct into the \textbf{Hard} class. For those examples that only require one-step processing, we categorize them into the \textbf{Easy} class. The rest of examples are categorized into \textbf{Medium} class.
As a result, the proportion of examples becomes 30\%:50\%:20\% for Easy:Medium:Hard.
Please refer to Table~\ref{tab:task_difficulty} for more details in Appendix~\ref{app:task-categ}.
As for the performance with respect to task difficulty, ReAct achieves the success rates of 34.0\%, 5.0\%, and 0.0\% on the Easy, Medium, and Hard examples, respectively. Our CoAct improves these values to 52.0\%, 16.0\%, and 0.0\%, accordingly.

\paragraph{Error analysis on the medium-difficulty examples.}

We especially investigate the failure cases in the ``Product Information Retrieval'' class, in order to uncover valuable insights to further improving the models. Below are our findings:
\begin{itemize}[leftmargin=10pt]
    \item \textit{Planning Inadequacies}: About 40\% of CoAct's failures are attributed to planning inadequacies stemming from deficiencies in the global planning agent. This category highlights errors arising from an insufficient understanding of the task, leading to inaccuracies in the initial global plan.
    The primary conclusion is the imperative \textbf{integration of web page-specific knowledge} into CoAct's planning process. Future efforts will prioritize enhancing the model's comprehension of task requirements through knowledge retrieval. This approach ensures a nuanced understanding of the web page's structure and content, mitigating planning inadequacies.
    \item \textit{Iterative and Repetitive Actions}: About 60\% of CoAct's failures involve iterative and repetitive actions, surpassing the maximum round limit for interaction between the global planning agent and the local execution agent. Mitigating this type of error necessitates optimizing the transfer process of plans by \textbf{introducing memory and experiential learning}. Incorporating memory mechanisms enables CoAct to learn from past interactions, reducing the occurrence of repetitive actions and enhancing overall efficiency.
\end{itemize}

\paragraph{Improving by integrating web page-specific knowledge from search engines. }

We conducted initial experiments to assess the impact of integrating web page-specific knowledge into our approach. 
Specifically, in the global planning process, we introduced a search step using search engines and augmented the text with brief passages not exceeding 100 words.
We evaluate this approach on two tasks, i.e., Shop and Gitlab, and report the results in Table~\ref{tab:search_engine_results}.
As seen, when enriched with information from search engine, CoAct is further improved by significant margins, namely, 24.0\% to 31.0\% on Shop and 10.0\% to 19.0\% on Gitlab. 
These results demonstrate the effectiveness of integrating web page-specific knowledge.  
Further investigation and fine-tuning are required to validate and optimize these findings for broader applications.

\begin{table}[t]
    \small
    \centering
    \begin{adjustbox}{max width=0.45\textwidth}
        \begin{tabular}{lccc}
            \toprule
            \bf Method & \bf Shop & \bf GitLab \\
            \midrule
            \textsc{CoAct w/ FS } & 24.0 & 10.0 \\
            ~~~~+ \textsc{Search Engine} & \bf 31.0 & \bf 19.0 \\
            \bottomrule
        \end{tabular}
    \end{adjustbox}
    \caption{Preliminary experiments on improving CoAct by integrating web page-specific knowledge from search engines. \textsc{+ Search Engine} includes additional information from specific web pages. 
    }
    \label{tab:search_engine_results}
\end{table}

%% file: latex/4-conclusion.tex
\section{Conclusion}

In this work, we propose the CoAct framework, composed of a global planning agent and a local execution agent, which transfers the hierarchical planning and collaboration patterns in human society to LLM systems. 
Experimental results on the WebArena benchmark show that CoAct can re-arrange the process trajectory when facing failures, and achieves superior performance over ReAct on long-horizon web tasks. 

%% file: latex/5-limitations.tex
\section{Limitations}

Despite representing a significant advancement in multi-agent collaboration for task execution, CoAct exhibits several notable limitations uncovered by our research:

\textbf{Planning Inadequacies:} Approximately 40\% of CoAct's failures stem from deficiencies in the global planning agent, leading to inaccuracies in initial plan formulation. We believe, enhancing CoAct's planning process with domain-specific knowledge could bolster task comprehension and robustness.
    
\textbf{Iterative and Repetitive Actions:}  Approximately 60\% of CoAct's failures stem from deficiencies in the global planning agent, resulting in inaccuracies in initial plan formulation. We have not implemented an efficient memory mechanism to address these issues, potentially limiting improvements in operational efficiency.
    
\textbf{Integration of Web Page-Specific Knowledge:} Initial experiments reveal promising outcomes in integrating web page-specific knowledge, yielding significant performance improvements. However, further refinement is necessary to generalize these findings across diverse application contexts.

These identified limitations underscore critical avenues for future research and enhancement, such as refining knowledge integration, optimizing interaction protocols, and safeguarding data integrity in training datasets.

%% file: latex/6-ethics.tex
\section{Ethical Considerations}

The development of advanced autonomous agents raises significant ethical considerations that must be carefully addressed. Key concerns include ensuring fairness and inclusivity to prevent discrimination, implementing robust safety measures to mitigate potential harms, ensuring transparency in decision-making processes for accountability and trustworthiness, and considering the implications of multi-agent interactions. This research adheres to the highest ethical standards and best practices by exclusively utilizing publicly accessible datasets, thereby avoiding any use of proprietary or confidential information and ensuring its ethical integrity.

%% file: latex/algorithm.tex
\begin{algorithm}
\caption{CoAct Framework}
\small
\begin{enumerate}
    \setlength\itemsep{0pt}
    \setlength\parskip{0pt}
    \item[\textbf{Input:}] Task $T$, Planner $GP$, Agents $LA$
    \item[\textbf{Output:}] Completed task and validation summary
    \item Initialize $GP$ and $LA$ for CoAct
    \item Delegate $T$ to $GP$
    \item $Plan_{g} \leftarrow GP(T)$
    \item for $\text{phase} = 1$ to $|Plan_{g}|$ do
    \item \quad $Subtask_{g} \leftarrow Plan_{g}[\text{phase}]$ 
    \item \quad for $i = 1$ to $|LA|$ do
    \item \quad \quad $Subtask_{l} \leftarrow Subtask_{g}[i]$ 
    \item \quad \quad $A_{l} \leftarrow LA[i].A(Subtask_{l})$
    \item \quad \quad $V_{l} \leftarrow LA[i].validate(A_{l})$ 
    \item \quad \quad if $\neg V_{l}$ then
    \item \quad \quad \quad $Replan_{g} \leftarrow GP.replan(Subtask_{g})$ 
    \item \quad \quad \quad if $Replan_{g}$ accepted then
    \item \quad \quad \quad \quad $Plan_{g} \leftarrow Replan_{g}$ 
    \item \quad \quad \quad end if
    \item \quad \quad end if
    \item \quad end for
    \item end for
    \item $T_{c} \leftarrow LA.executeTasks()$ 
    \item $V_{g} \leftarrow GP.validate(T_{c})$ 
    \item \textbf{Return} $T_{c}$, $V_{g}$
\end{enumerate}
\label{algo:coact}
\end{algorithm}

%% file: latex/global-prompt.tex
\begin{table*}[t]
    \small
    \centering
    \begin{adjustbox}{max width=\textwidth}
\begin{tabular}{|m{0.9\textwidth}|}
\hline
\textbf{Introduction} \\
\hline
You are an autonomous intelligent agent playing the role of a strategic leader in a multi-tier task execution structure, tasked with navigating a web browser. You will be given web-based tasks. Your responsibility is to provide high-level, strategic plans that can be broken down into smaller tasks by the local agent.

\textbf{Information Available:}
\begin{itemize}
    \item The user's objective: This is the task you're trying to complete.
    \item The current web page's accessibility tree: This is a simplified representation of the webpage, providing key information.
    \item The current web page's URL: This is the page you're currently navigating.
    \item The open tabs: These are the tabs you have open.
\end{itemize}

\textbf{Actions You Can Perform:}
\begin{itemize}
    \item \texttt{global plan}: Construct a multi-stage global plan, providing separate subtask descriptions and expected states for each phase.
    \item \texttt{decide}: Based on the description in the request for a new global plan submitted by the local agent, decide whether to agree to the re-planning. If so, the next action is to revise; if not, the next action is to overrule.
    \item \texttt{revise}: Facing the local agent's request, you chose to revise your previously made global plan. Please reconsider the characteristics of the task based on the description in the request and make a new global plan.
    \item \texttt{overrule}: Facing the local agent's request, you believe your previous global plan is correct, so you refuse to adjust it and overrule the local agent's request. You believe they should adjust their local plan instead. Please give them suggestions for modifications.
    \item \texttt{collation}: Collation the final result to meet the need of the task.
\end{itemize}

\textbf{Meta Prompts for Each Action:}
\begin{itemize}
    \item \texttt{global\_plan}: Your role is to construct a multi-stage global plan, providing separate subtask descriptions and expected states for each phase. Ensure that your plan is comprehensive and covers all aspects of the task.
    \item \texttt{decide}: The local agent encountered issues with the global planner's global plan, and he believes it's necessary to replan globally. Here are the reasons he proposed: \{reasons\}. Your current task is to decide whether to agree to the re-planning based on the description in the request for a new global plan submitted by the local agent. If you agree, the next action is to \texttt{revise}; if not, the next action is to \texttt{overrule}. Consider the implications of your decision and provide clear reasoning for it. Make sure to give clear and constructive guidance for plan adjustments.
    \item \texttt{revise}: In response to the local agent's request, you have chosen to revise your previously made global plan. Reconsider the characteristics of the task based on the description in the request and create a new global plan. Ensure that your revised plan is well-detailed and addresses any issues identified.
    \item \texttt{collation}: Your role now involves collating the final result to meet the needs of the task. Ensure that the final output aligns with the global plan and that any necessary adjustments have been made. Provide a comprehensive summary of the task's completion.
\end{itemize}

\textbf{Prompt Template:}
\texttt{\textbackslash{}template}: OBSERVATION: \{observation\} \\
URL: \{url\} \\
OBJECTIVE: \{objective\} \\
PREVIOUS ACTION: \{previous\_action\} \\
\hline
\end{tabular}
    \end{adjustbox}
    \caption{Prompt for \textbf{global planning agent}.
    }
    \label{}
\end{table*}

%% file: latex/local-prompt.tex
\begin{table*}[t]
    \small
    \centering
    \begin{adjustbox}{max width=\textwidth}
\begin{tabular}{|p{0.9\textwidth}|}
\hline
\textbf{Introduction} \\
\hline
You are an autonomous intelligent agent playing the role of a subordinate employee responsible for local planning and execution of specific tasks in a multi-tier task execution structure, tasked with navigating a web browser. You will be given web-based tasks. The global agent has set a global plan for the tasks, divided into multiple phases. These phase plans will be given to you one by one. Your responsibility is to dissect the present phase's subtask into a detailed sequence of Page Operation Actions.

\textbf{Information Available:}
\begin{itemize}
    \item The objective of the current phase: This is the task you're trying to complete now.
    \item The current web page's accessibility tree: This is a simplified representation of the webpage, providing key information.
    \item The current web page's URL: This is the page you're currently navigating.
    \item The open tabs: These are the tabs you have open.
    \item The previous action: This is the action you just performed. It may be helpful to track your progress.
\end{itemize}

\textbf{Actions You Can Perform:}
[...same as code in WebArena]

\textbf{Meta Prompts for Each Action:}
\begin{itemize}
    \item \texttt{local\_plan}: Your objective now is to complete a specific task on the current webpage. Analyze the accessibility tree and page content carefully. Consider using Page Operation Actions to interact with elements. Follow the examples and provide a clear sequence of actions to accomplish the task. Please adhere to the following output template:
    \begin{verbatim}
    **Action 1:** [action]
    **Action 2:** [action]
    ...
    **Action m:** [action]
    \end{verbatim}
    \item \texttt{pass\_check}: Now, your role is to ensure the successful execution of actions in the global plan. Verify the results of these actions and compare them to the global plan. If they align, proceed to the next phase and output the action decision as ```move```. If discrepancies arise, you have two options: 1) If you suspect issues with your local plan, output the action ```revise```, or 2) If you suspect problems with the global planner's plan, trigger a request for replanning by outputting the action ```request```. 
    If the actions align with the global plan, explain the reasons for this alignment. If discrepancies arise, provide detailed reasons for your action decision:
    \begin{verbatim}
    Action: [action]
    Reasons: [reasons]
    \end{verbatim}
    \item \texttt{false\_check}: You have encountered an exception in the execution process. Your current responsibility is to meticulously inspect the execution results of actions and identify the root causes of these exceptions. You have two options: 1) Suspect issues within your local plan and employ the action ```revise```, or 2) Suspect problems with the global planner's plan and trigger a request for replanning by executing the action ```request```.
    Provide detailed reasons for your action decision:
    \begin{verbatim}
    Action: [action]
    Reasons: [reasons]
    \end{verbatim}
    \item \texttt{revise}: Now, you have analyzed the situation and decided adjustments are needed to the local plan. Here are the reasons you proposed: {reasons}. Provide a revised plan using Page Operation Actions, and make sure to follow the format for action generation as mentioned earlier.
    \item \texttt{overruled}: Facing your request, the global planner believes his previous global plan is correct and refuses to adjust it and overrules your request. Here are the reasons he proposed: {reasons}. Based on this information and your past experience, provide a revised plan using Page Operation Actions, and make sure to follow the format for action generation as mentioned earlier.
\end{itemize}

\textbf{Prompt Template:}
\texttt{\textbackslash{}template}: OBSERVATION: \{observation\} \\
URL: \{url\} \\
OBJECTIVE: \{objective\} \\
PREVIOUS ACTION: \{previous\_action\} \\
\hline
\end{tabular}
    \end{adjustbox}
    \caption{Prompt for \textbf{local execution agent}.
    }
    \label{}
\end{table*}